%% file: main.tex
\pdfoutput=1

\documentclass[11pt]{article}

\usepackage[]{EMNLP2022}

\usepackage{times}
\usepackage{latexsym}

\usepackage[T1]{fontenc}

\usepackage[utf8]{inputenc}

\usepackage{microtype}

\usepackage{graphicx}
\usepackage{amsmath}
\usepackage{amsfonts}
\usepackage{array,booktabs,arydshln,xcolor}
\usepackage{graphics}
 \usepackage[most]{tcolorbox}
\usepackage{lipsum}
\usepackage{pifont}
\usepackage{hyperref}
\usepackage{makecell}
\usepackage[super]{nth}
\usepackage{wasysym}
\usepackage{float}
\usepackage{resizegather}
\usepackage{adjustbox}
\usepackage{booktabs}
\usepackage{multirow}
\usepackage{makecell}
\usepackage{subcaption}

\usepackage{inconsolata}

\title{Towards Few-Shot Identification of Morality Frames using In-Context Learning}

\author{Shamik Roy, Nishanth Sridhar Nakshatri, Dan Goldwasser \\
  Department of Computer Science \\
  Purdue University\\
  West Lafayette, IN, USA\\
  \texttt{\{roy98, nnakshat, dgoldwas\}@purdue.edu} }

\begin{document}
\maketitle
\begin{abstract}
Data scarcity is a common problem in NLP, especially when the annotation pertains to nuanced socio-linguistic concepts that require specialized knowledge. As a result, few-shot identification of these concepts is desirable. Few-shot in-context learning using pre-trained Large Language Models (LLMs) has been recently applied successfully in many NLP tasks. In this paper, we study few-shot identification of a psycho-linguistic concept, Morality Frames \citep{roy2021identifying}, using LLMs. Morality frames are a representation framework that provides a holistic view of the moral sentiment expressed in text, identifying the relevant moral foundation~\cite{haidt2007morality} and  at a finer level of granularity, the moral sentiment expressed towards the entities mentioned in the text. Previous studies relied on human annotation to identify morality frames in text which is expensive. In this paper, we propose prompting based approaches using pretrained Large Language Models for identification of morality frames, relying only on few-shot exemplars. We compare our models' performance with few-shot RoBERTa and found promising results.
\end{abstract}

\input{1introduction}
\input{2related_works}
\input{3dataset}
\input{4model}
\input{5experiment}
\input{6summary_future_works}

\input{limitations_ethical_statement}

\bibliography{anthology,custom}
\bibliographystyle{acl_natbib}

\appendix

\input{7appendix}

\end{document}

%% file: 1introduction.tex
\section{Introduction}\label{sec:introduction}
While the NLP field has seen tremendous progress over the last decade, building models capable of identifying abstract concepts remain a highly challenging problem. This difficulty stems from two key reasons. First, these concepts can manifest in very different ways in text. For example, the concept of \textit{fairness}, that we discuss at length in this paper, can be discussed in the context of the abortion debate (e.g., \textit{``right to privacy''}) or in the context of Covid-19 vaccination (e.g., \textit{``everyone should have access to the vaccine''}). Learning to identify instances of this concept in previously unseen contexts remains a challenge.  Second, building NLP models using the supervised learning paradigm requires humans to annotate data, which for such tasks is a cognitively demanding process. In this paper, we investigate whether the recently introduced paradigm of zero/few shot learning using Large Language Models~\cite{brown2020language} is better equipped to deal with these challenges. We focus on a recently introduced framework for analyzing moral sentiment, called \textit{morality frames}~\cite{roy2021identifying}. This framework builds on, and extends, moral foundation theory~\cite{haidt2007morality}, which identifies five moral values (i.e., foundations, each with a positive and a negative polarity) central to human moral sentiment which include Care/Harm, Fairness/Cheating, Loyalty/Betrayal, Authority/Subversion, and Purity/Degradation. Morality frames is a relational framework that identifies expressions of the moral foundations in text and associates moral roles with entities mentioned in it (see Section~\ref{sec:dataset} for details). 

Unlike previous approaches to this task~\cite{roy2021identifying,pacheco-etal-2022-holistic} which use annotated data to train a relational classifier using DRaiL~\cite{pacheo-etal-DRAIL}, we define the task as a zero/few shot problem. We rely on in-context learning using Large Language Models for the identification of morality frames. In in-context learning, a desired NLP task is framed as a text generation problem where the Large Language Models are provided with zero/few shot input-output pairs and prompted to generate label for the test data point without updating parameters of the LLMs \cite{min2021recent}. 

In this paper, we introduce several prompting techniques for LLMs for the identification of morality frames in tweets that rely on only few-shot examples. We compare our models' performance with few-shot RoBERTa-based \cite{liu2019roberta} classifiers. We found that prompting-based techniques underperform RoBERTa in identification of subtle concepts like moral foundations, but in case of moral role identification, the prompting-based techniques outperforms RoBERTa by a large margin. Note that moral roles are directed towards entities and are more evident than subtle moral foundations. 

Our promising findings in this paper suggest that in-context learning approaches can be useful in many Computational Social Science related tasks and we propose a few potential future directions of this work.



%% file: 2related_works.tex
\section{Related Works}\label{sec:related-works}
There has been a lot of work towards exploiting existing knowledge in pretrained Large Language Models (LLMs) and improving its few-shot abilities on various downstream tasks in NLP. Some of these works have been influenced from areas related to instruction-based NLP~\cite{goldwasser2014learning}.  ~\citeauthor{mishra2021cross}, \citeyear{mishra2021cross} fine-tuned a 140M parameter BART~\cite{lewis2019bart} model using instructions and few-shot examples for various NLP tasks such as text classification, question answering, and text modification. This work suggests that augmenting instructions in the fine-tuning process improves model performance on unseen tasks. On similar lines, through a large scale experiment with over 60 different datasets, ~\citeauthor{wei2021finetuned},~\citeyear{wei2021finetuned} showed that instruction tuning on a LLM ($\approx$137B parameters) improves zero and few-shot capabilities of these models. Other notable works~\cite{min2021metaicl,sanh2021multitask} show that even a relatively smaller language model can achieve substantial improvement in a similar setting. Furthermore, ~\citeauthor{schick2020exploiting},~\citeyear{schick2020exploiting} use cloze-style phrases in a semi-supervised manner to help LM assign a sentiment label for the text classification task. 

Another line of work focuses on improving LM on downstream tasks with no parameter updates.~\citeauthor{brown2020language}, \citeyear{brown2020language} proposed to improve LLM few-shot performance by conditioning on concatenation of training examples without any gradient updates. Other works~\cite{min2021noisy,zhao2021calibrate} have further improved this work and have shown consistent gains in various NLP tasks. In addition,~\citeauthor{wei2022chain}, \citeyear{wei2022chain} shows that sufficiently large LM can exploit its innate reasoning abilities to solve complex tasks when provided with a series of intermediate steps during prompting.

However, having a generalized LLM may have poor performance when the downstream task needs nuanced understanding of the text or is very different from language modeling in nature. While    
~\citeauthor{schick2020exploiting},  \citeyear{schick2020exploiting} and ~\citeauthor{gao2020making}, \citeyear{gao2020making} have studied sentiment classification task in few-shot settings, not many works are available towards utilizing LLM without finetuning it to understand more nuanced concepts like political framing \cite{boydstun2014tracking}, moral foundations \cite{haidt2004intuitive, haidt2007morality}, among others. 

Previous work \cite{royweakly} has performed nuanced analysis of political framing by breaking the policy frames proposed by \citeauthor{boydstun2014tracking}, \citeyear{boydstun2014tracking}, into fine-grained sub-frames. It was observed that the sub-frames better captured political polarization by providing a structural breakdown of policy frames. A later work \cite{roy2021analysis} studied the Moral Foundation Theory \cite{haidt2004intuitive, haidt2007morality} at entity level and proposed a knowledge representation framework for organizing moral attitudes directed at different entities. The structured framework is named morality frames \cite{roy2021identifying}. These nuanced structural frameworks, such as, frames, sub-frames, entity-centric moral sentiments (morality frames), are expensive to annotate as they largely depend on human knowledge. A few-shot automatic identification of such concepts is required to save manual human-effort and for performing these studies at scale. In this paper, we take the first step towards the analysis on how well LLMs can understand these psycho-linguistic concepts in few-shot settings. As our first study, we explore in-context learning of morality frames in this paper and leave the study of framing and sub-frames as a future work.

%% file: 3dataset.tex
\section{Dataset}\label{sec:dataset}
We conduct our study on the dataset proposed by \citet{roy2021identifying}. In this dataset, there are $1599$ political tweets from US politicians that are annotated for moral foundations by \citet{johnson2018classification}. \citet{roy2021identifying} proposed Morality Frames and broke down the sentence level moral foundations into nuanced moral role dimensions that capture sentiment towards entities expressed in the text. The moral foundations and corresponding moral roles can be found in Table \ref{tab:moral-foundation-roles}. \citet{roy2021identifying} annotated the dataset proposed by \citet{johnson2018classification} for these moral sentiments towards entities.

\begin{table}[ht!]
\begin{center}
 \scalebox{0.70}{\begin{tabular}{>{\arraybackslash}m{5.5cm}|>{\arraybackslash}m{4.5cm}} 
 \hline 
 
\textbf{Moral Foundations} & \textbf{Moral Roles}\\ [0.5ex]
\hline
\textbf{Care/Harm:} Care for others, generosity, compassion, ability to feel pain of others, sensitivity to suffering of others, prohibiting actions that harm others. & 
\makecell[l]{Target of care/harm\\Entity causing harm\\Entity providing care}\\
\hline
\textbf{Fairness/Cheating:} Fairness, justice, reciprocity, reciprocal altruism, rights, autonomy, equality, proportionality, prohibiting cheating. & 
\makecell[l]{Target of fairness/cheating\\Entity ensuring fairness\\Entity doing cheating}\\
\hline
\textbf{Loyalty/Betrayal:} Group affiliation and solidarity, virtues of patriotism, self-sacrifice for the group, prohibiting betrayal of one’s group. & 
\makecell[l]{Target of loyalty/betrayal\\Entity being loyal\\Entity doing betrayal}\\
\hline
\textbf{Authority/Subversion:} Fulfilling social roles, submitting to authority, respect for social hierarchy/traditions, leadership, prohibiting rebellion against authority. & 
\makecell[l]{Justified authority\\Justified authority over\\Failing authority\\Failing authority over}\\
\hline
\textbf{Purity/Degradation:} Associations with the sacred and holy, disgust, contamination, religious notions which guide how to live, prohibiting violating the sacred. & 
\makecell[l]{Target of purity/degradation\\Entity preserving purity\\Entity causing degradation}\\
\hline
\end{tabular}}
\caption{Morality Frames: Moral foundations and their associated roles. (Adopted from \cite{roy2021identifying}).}
\label{tab:moral-foundation-roles}
\end{center}
\end{table}

In this paper, our goal is to study the identification of morality frames when only few-shot training examples are available. To build this setup, we randomly sampled $10$ tweets from each of the $5$ moral foundations, and used it as training set. We use Large Language Models (LLMs) for in-context learning that are expensive and resource heavy even for inference only. So, we benchmark our approaches using a smaller test set containing randomly sampled $20$ tweets per moral foundation. It resulted in $103$ and $207$ tweet-entity pairs in the training and the test set, respectively.  

%% file: 4model.tex
\section{Task Definition}\label{sec:task-definition}
The identification of morality frame in a tweet involves the following two steps.\\

\noindent\textbf{Identification of Moral Foundation:} Given a tweet text $t$, the task is to identify the moral foundation expressed in the tweet.\\

\noindent\textbf{Identification of Moral Roles of Entities:} After identification of moral foundation, the second step is to identify the moral roles of entities in the tweet. We study this step in the following two settings.  

\begin{itemize}
    \item \textbf{Entities are pre-identified:} In this setting, the assumption is that the entities are already identified in the tweet text. The task is to assign moral roles to them. So, given a tweet $t$, an entity $e$ mentioned in the tweet, and the moral foundation label of the tweet $m$, the task is to identify the moral role of $e$ in $t$.
    
    \item \textbf{Entities are not pre-identified:} In this setting, a tweet $t$, and its corresponding moral foundation label $m$ is known in prior. The task is to identify the entities mentioned in the tweet, and their corresponding moral roles.
\end{itemize}

\begin{figure}[h!]
  \includegraphics[width=0.5\textwidth]{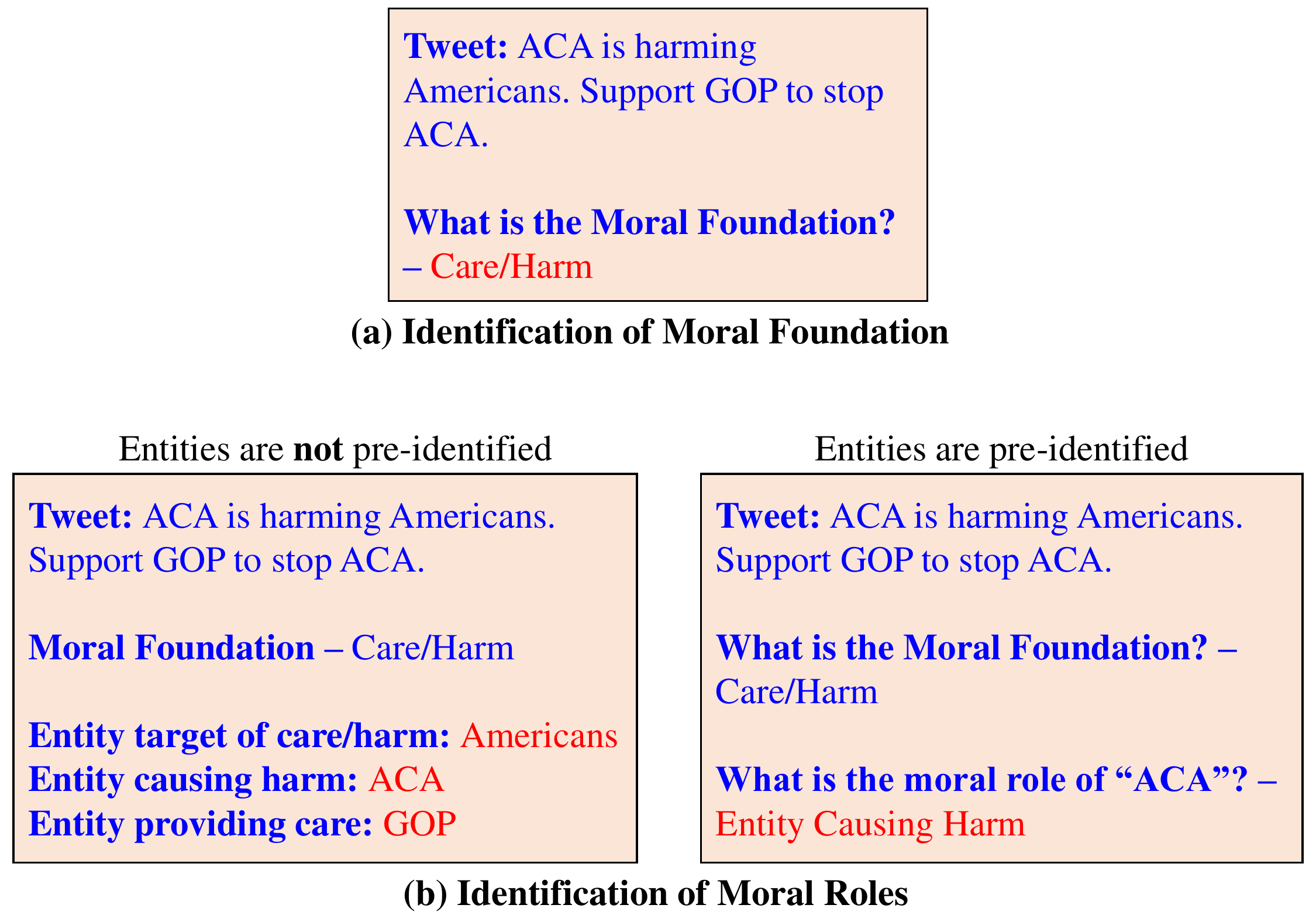}
  \caption{Morality frames identification task. Input for each step is colored in blue and expected outputs are colored in red.}
  \label{fig:settings}
\end{figure}

Examples of the tasks can be found in Figure \ref{fig:settings}.

\section{Few-Shot Identification of Morality Frames using Large Language Models}\label{sec:model}

\subsection{In-Context Learning} In-context learning using pretrained LLMs has been shown effective in few-shot scenarios in previous studies for different NLP tasks \citep{brown2020language,wei2022chain,reif2021recipe}. LLMs are pretrained on huge amount of web-crawl, books and Wikipedia text. Hence, they are expected to carry world-knowledge. As a result, they are able to perform many NLP tasks using only few-shot training examples without any further fine-tuning or gradient updates. In the in-context learning paradigm, the downstream task is framed as a text generation problem and the model is prompted to generate the next tokens \cite{min2021recent}. These tokens are mapped to desired output labels in classification tasks. In this work, we assume that only few-shot examples are given for the morality frames identification task. So, we apply in-context learning approach for this purpose to perform different steps of the task defined in Section \ref{sec:task-definition}. Note that we do not update LLM parameters in this process. The proposed in-context learning approaches are described in the subsequent sections.

\subsection{Moral Foundation (MF) Identification}
Following the previous works, we frame the task of moral foundation identification as a text generation problem where the model is prompted to generate the moral foundation label of a tweet. To this end, we experiment with two different types of prompting techniques.\\

\begin{figure}[h!]
  \centering
  \includegraphics[width=0.45\textwidth]{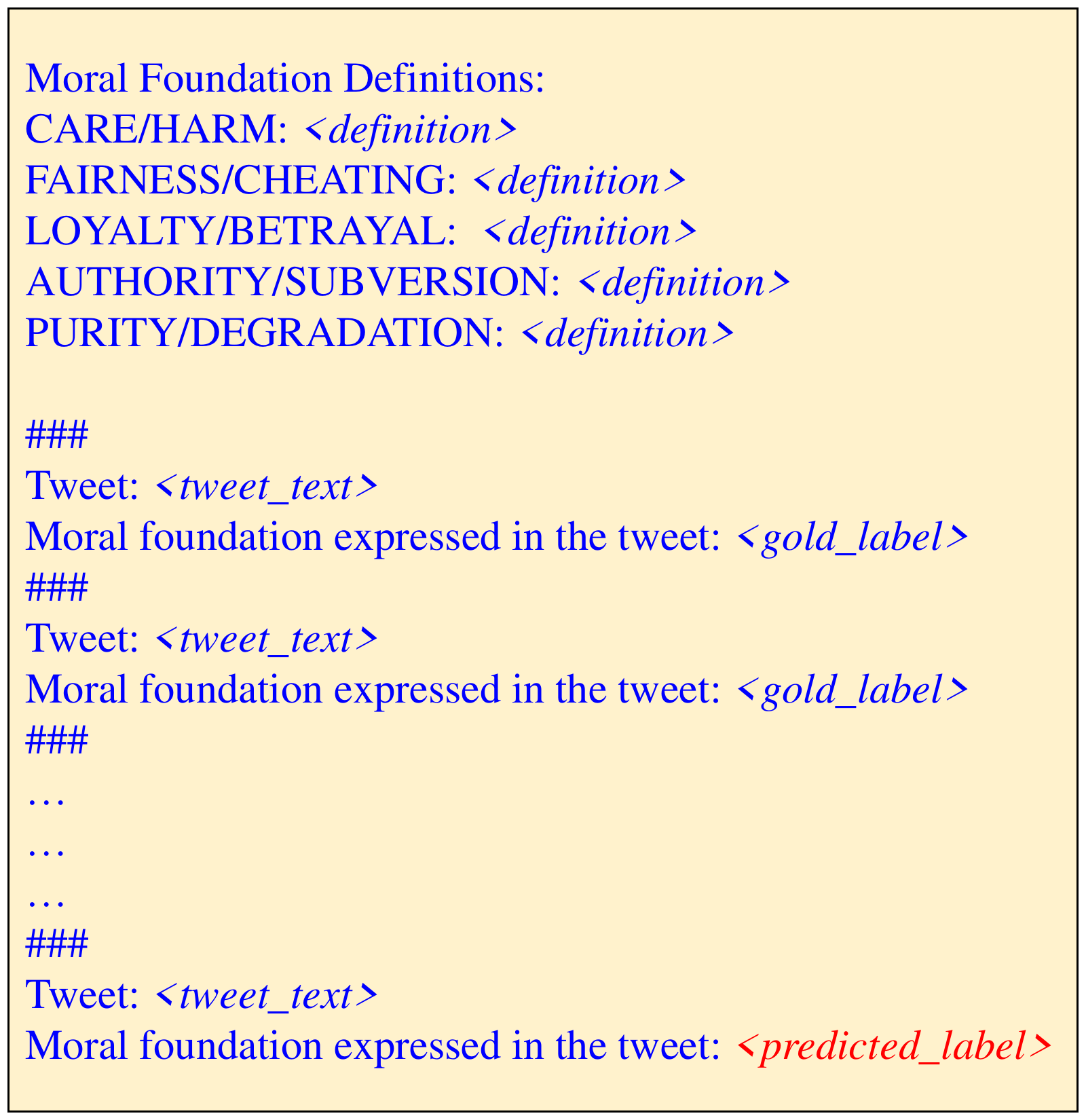}
  \caption{Prompt template for identification of moral foundation in one pass. The blue colored segment is input prompt and the red colored segment is the generated output by the LLMs. Example of this prompt template can be seen in Appendix \ref{appx:prompt-examples}: Figure \ref{fig:mf-direct-example}.}
  \label{fig:mf-direct-template}
\end{figure}

\noindent\textbf{MF identification in one pass:} In this method, we provide the moral foundation definitions (from Table \ref{tab:moral-foundation-roles}) in the beginning of the prompt as a guideline for the language model. Then, few-shot training examples and their associated labels are provided in the prompt. Finally, the test tweet is provided as the last example in the prompt and the model is expected to generate the moral foundation label of this tweet. The prompt template for this approach can be seen in Figure \ref{fig:mf-direct-template}.\\ 

\begin{figure}[t!]
    \begin{center}
    \begin{subfigure}[t]{0.45\textwidth}
        \centering
        \includegraphics[width=1\textwidth]{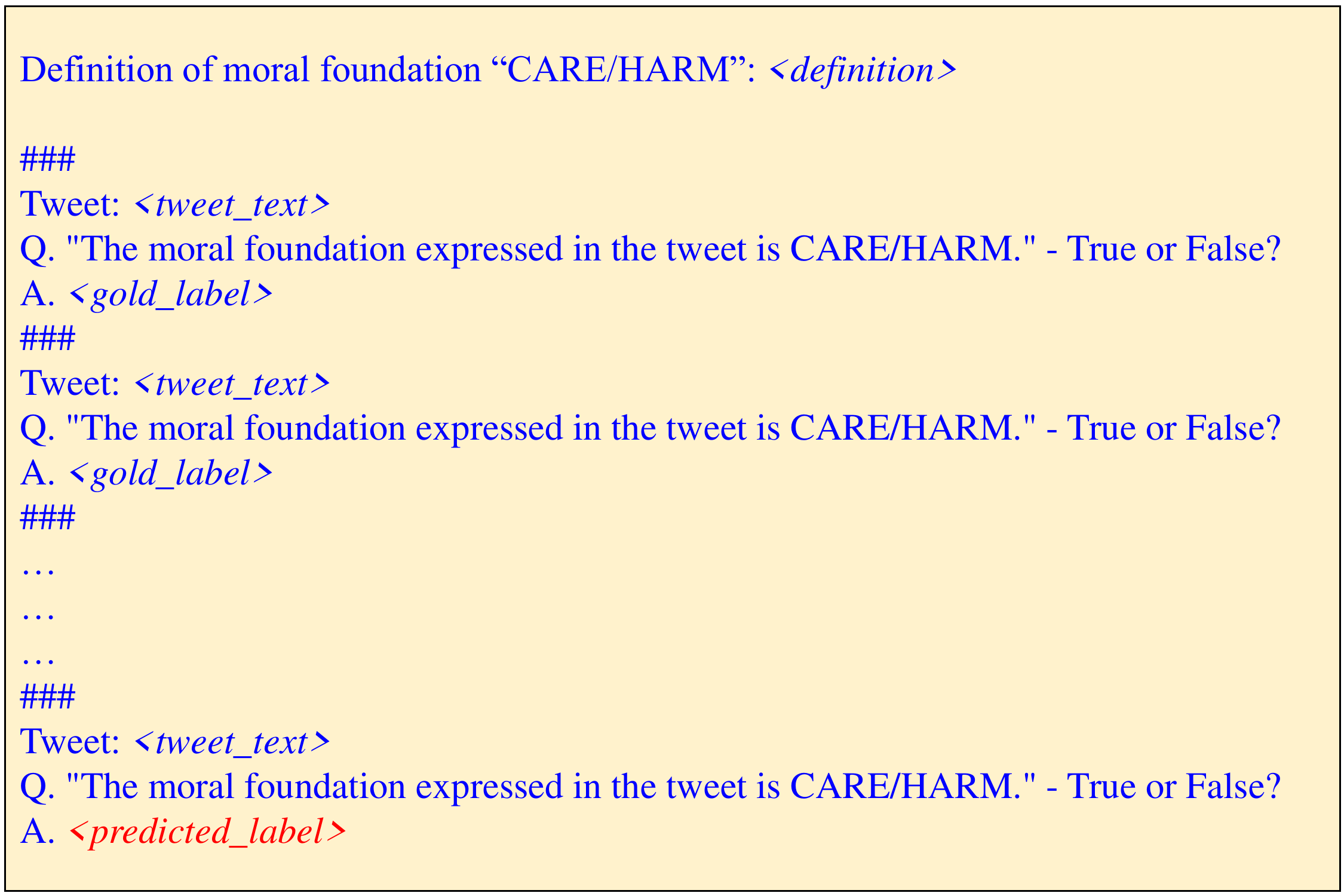}
        \caption{Prompt template for one-vs-all MF identification in case of `Care/Harm'.}
        \label{fig:step-1}
    \end{subfigure}
    \\
    \begin{subfigure}[t]{0.45\textwidth}
        \centering
        \includegraphics[width=1\textwidth]{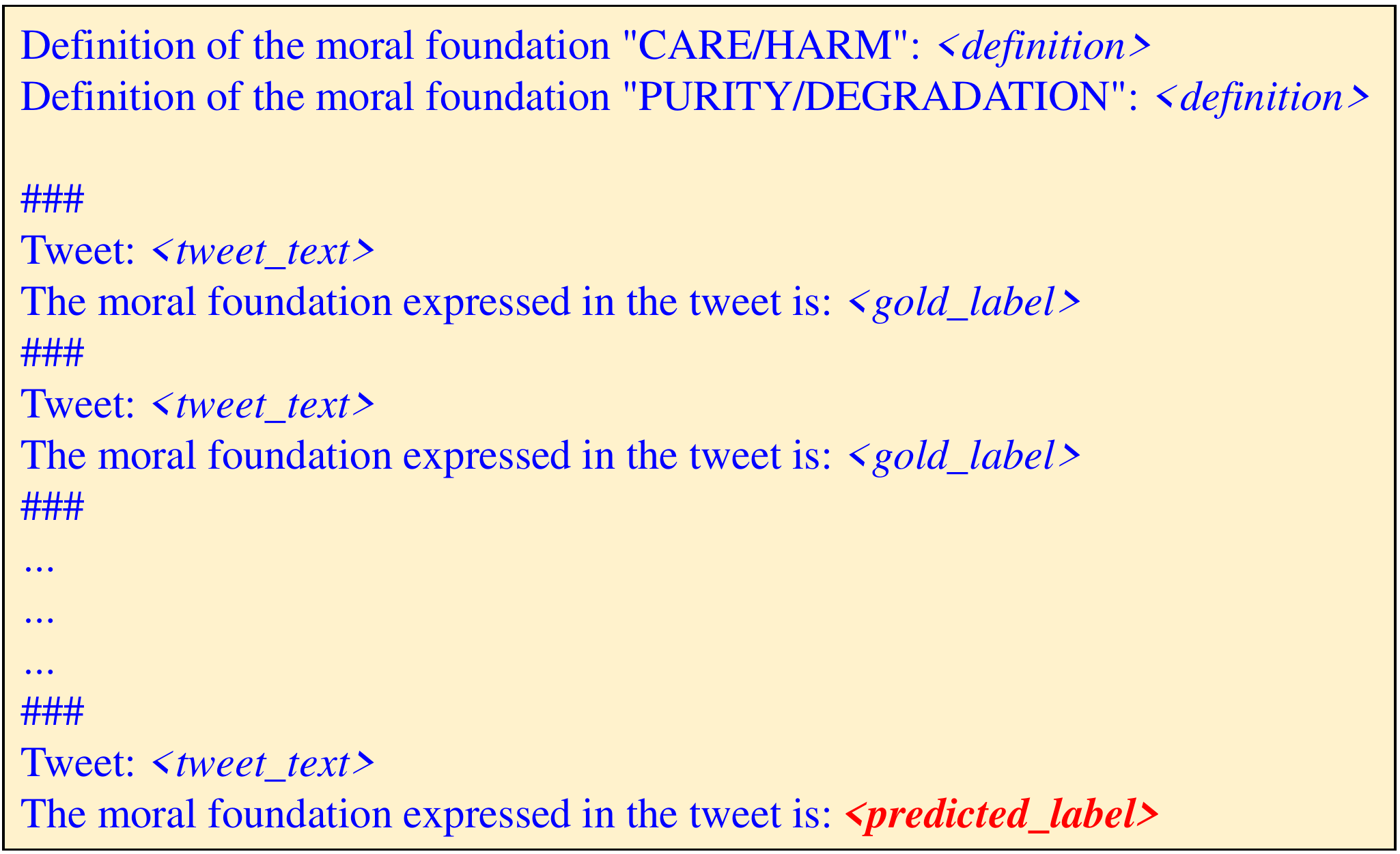}
        \caption{Prompt for tie-breaking between two MFs. For example, between `Care/Harm' and `Purity/Degradation'.}
        \label{fig:step-2}
    \end{subfigure}
    \caption{Prompt templates for moral foundation identification technique in one-vs-all manner. The blue colored segments are input prompts and the red colored segments are the generated output by the LLMs. Corresponding prompt example can be seen in Appendix \ref{appx:prompt-examples}: Figure \ref{fig:mf-one-vs-all-example}.}
    \label{fig:mf-one-vs-all}
    \end{center}
\end{figure}

\noindent\textbf{MF identification in one-vs-all manner:} Identification of moral foundations in one-pass might be difficult for the language models. So, we propose one-vs-all prompting approach where the language model is prompted to predict if a certain moral foundation is present in the tweet. This step is repeated for each of the five moral foundations. The moral foundation predicted with the highest confidence is consolidated as the predicted label. To obtain the confidence score, we prompt the language model multiple times with different random seeds to generate multiple predictions for a single tweet. The final confidence score is the percentage of times a specific moral foundation is generated by the LLM. In case there is a tie between two moral foundation labels, we perform a second prompting step, where few-shot prompting enables to break the tie between moral foundations.\footnote{In case of tie among more than two moral foundations, we break that by randomly selecting one.} Prompt templates for these two steps can be seen in Figure \ref{fig:mf-one-vs-all}.

\subsection{Moral Role Identification of a Pre-identified Entity}
Post prediction of the moral foundation label, the next step is to identify moral roles of entities as described in the Section \ref{sec:task-definition}. Given a test tweet, and a predicted moral foundation label for it, we prompt the LLMs to generate moral role of an entity in a tweet only from the associated moral roles to the predicted moral foundation. For example, if a tweet is identified to be having the moral foundation `Care/Harm', we prompt the language model to predict the the moral role of an entity mentioned in the tweet from only three moral roles that are associated to `Care/Harm', namely, `Entity target of care/harm', `Entity causing harm', `Entity providing care'. We propose two prompting approaches for this task.\\

\begin{figure}[h!]
  \centering
  \includegraphics[width=0.45\textwidth]{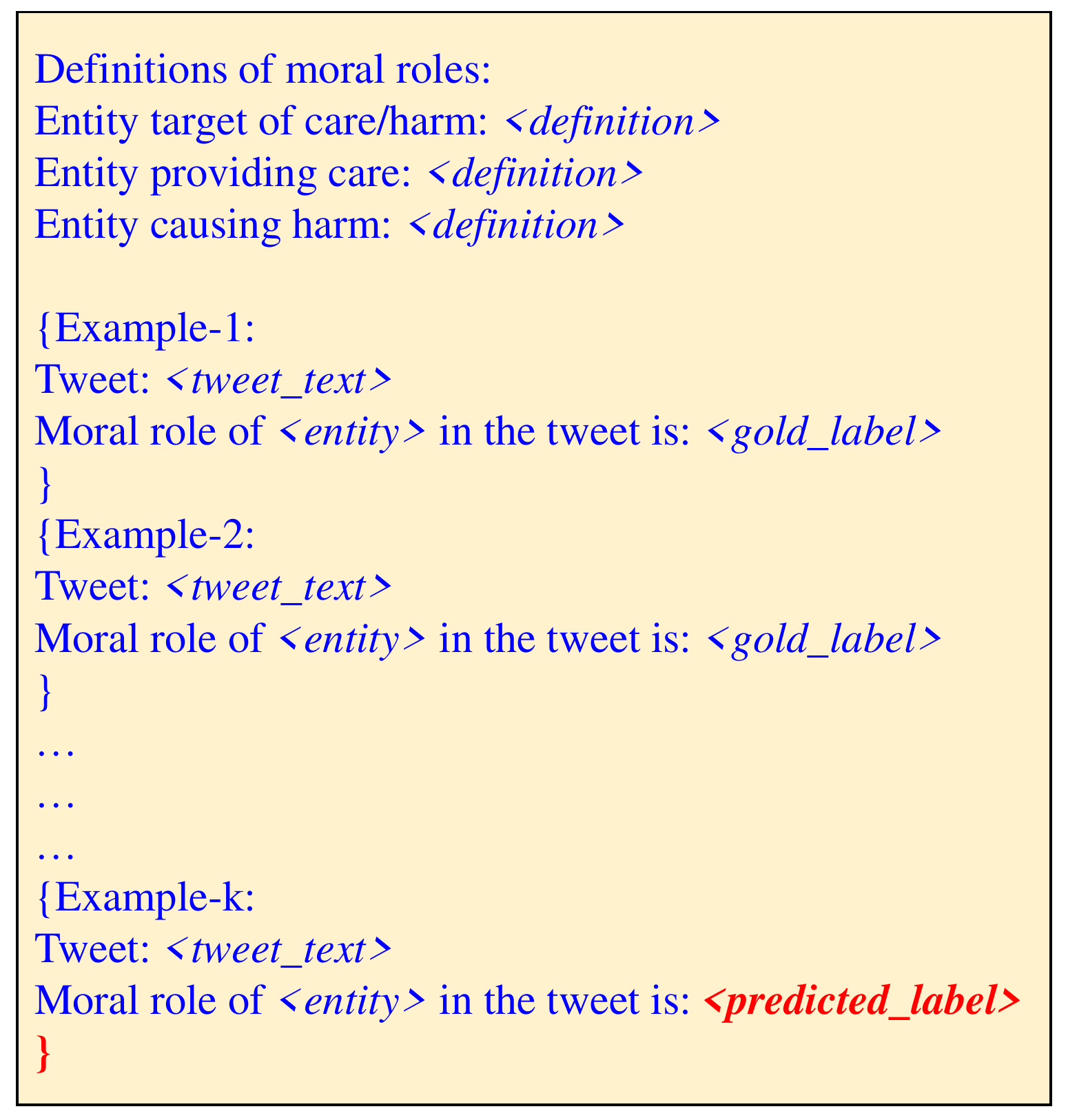}
  \caption{Prompt template for identification of moral role in one pass in case of `Care/Harm'. The blue colored segment is input prompt and the red colored segment is the generated output by the LLMs. Corresponding prompt example can be seen in Appendix \ref{appx:prompt-examples}: Figure \ref{fig:mf-role-direct-example}.}
  \label{fig:mf-role-direct-template}
\end{figure}

\noindent\textbf{Moral role identification in one pass:} We prompt the LLMs to directly identify moral role of a given entity from the corresponding moral roles in one pass using the prompt shown in Figure \ref{fig:mf-role-direct-template}. Following the moral foundation classification prompt template, we provide the description of the moral roles in the template as guideline. We come up with the definitions based on intuition.\\

\begin{figure}[t!]
    \begin{center}
    \begin{subfigure}[t]{0.45\textwidth}
        \centering
        \includegraphics[width=1\textwidth]{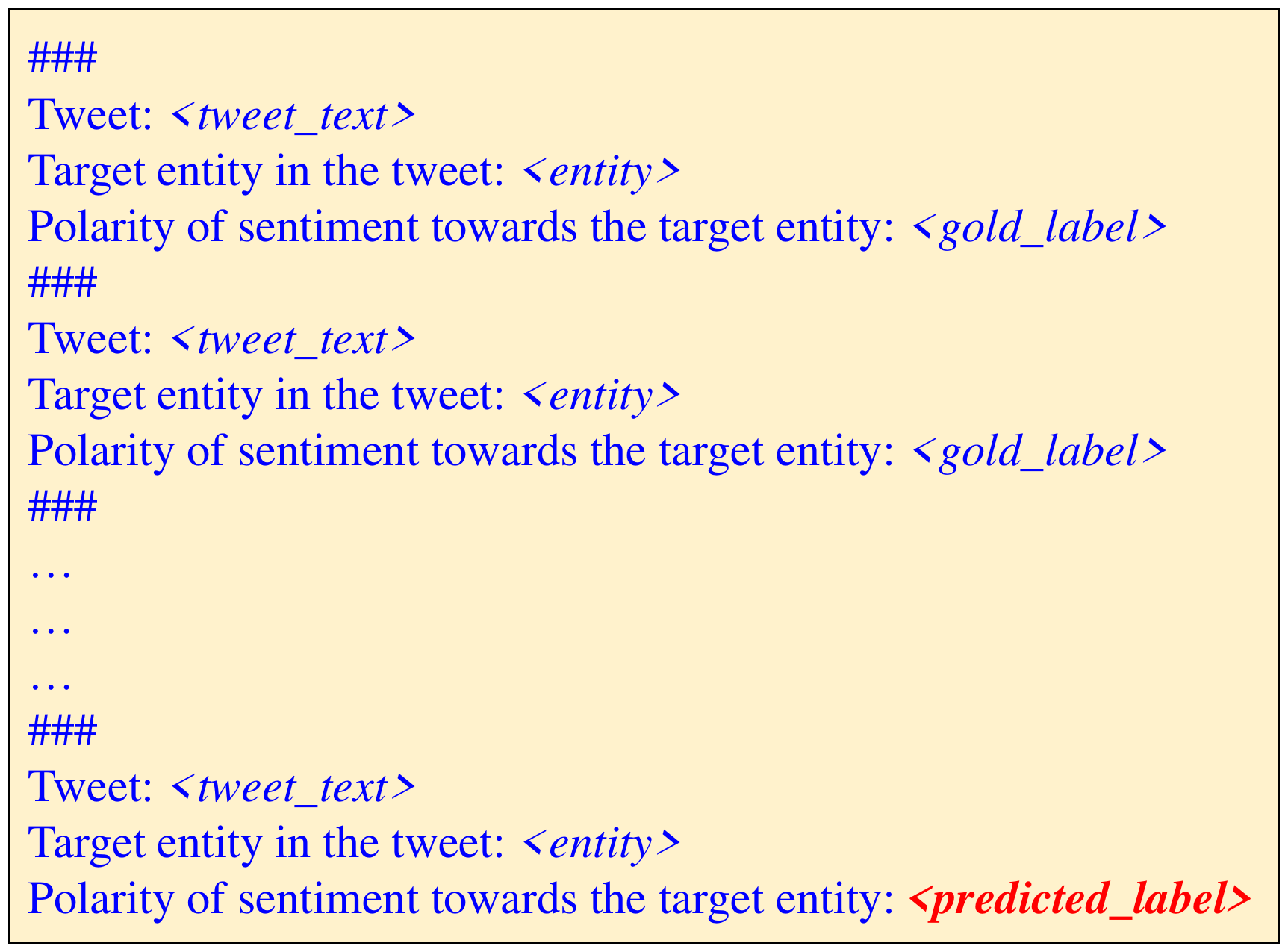}
        \caption{Step-1: Prompt template for identification of positive/negative sentiment towards entities.}
        \label{fig:mf-role-step-1}
    \end{subfigure}
    \\
    \begin{subfigure}[t]{0.45\textwidth}
        \centering
        \includegraphics[width=1\textwidth]{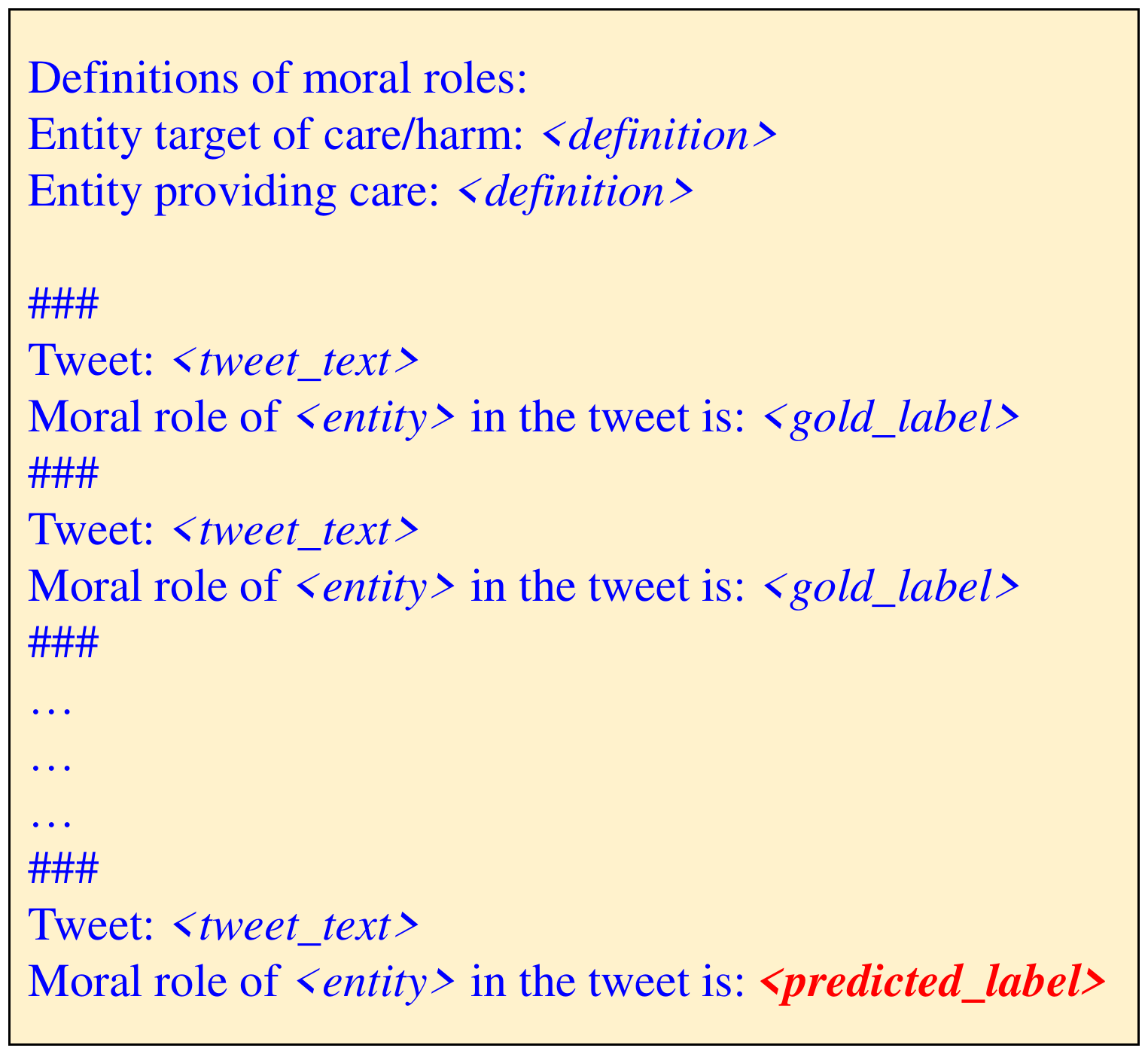}
        \caption{Step-2: Prompt template for differentiating among multiple positive moral roles in case of `Care/Harm'.}
        \label{fig:mf-role-step-2}
    \end{subfigure}
    \caption{Prompt templates for moral role identification by breaking the task in 2 steps. The blue colored segments are input prompts and the red colored segments are the generated output by the LLMs. Corresponding prompt examples can be seen in Appendix \ref{appx:prompt-examples}: Fig. \ref{fig:mf-role-two-steps-example}.}
    \label{fig:mf-role-two-steps}
    \end{center}
\end{figure}

\noindent\textbf{Moral role identification in two steps:} In the morality frames, different moral foundation roles intuitively carry either positive or negative sentiment towards them. For example, "entity causing harm", "entity violating fairness", "entity doing cheating", "failing authority" and "entity doing degradation" are the roles carrying negative sentiment towards them. The rest of the entity roles carry positive sentiment towards them. With this intuition, we break down the task of moral role identification in two steps. In the first step, we prompt the LLMs to identify the sentiment towards entities in "positive" and "negative" dimensions only by using the prompt structure in Figure \ref{fig:mf-role-step-1}. Now the entities discovered as having negative sentiment towards them directly maps to one of the five negative sentiments, each associated with only one of the moral foundations. Given the moral foundation is discovered in the previous step, we can readily map the entities with negative sentiments to one of the negative moral roles. Now, each moral foundation has two or more positive moral roles associated to them. To differentiate among them, we perform another prompting step where the LLMs are prompted to generate one of the positive moral roles for an entity in a tweet. The prompt template is shown in Figure \ref{fig:mf-role-step-2}.

\subsection{Identification of entities and corresponding moral roles jointly}
\begin{figure}[h!]
  \centering
  \includegraphics[width=0.45\textwidth]{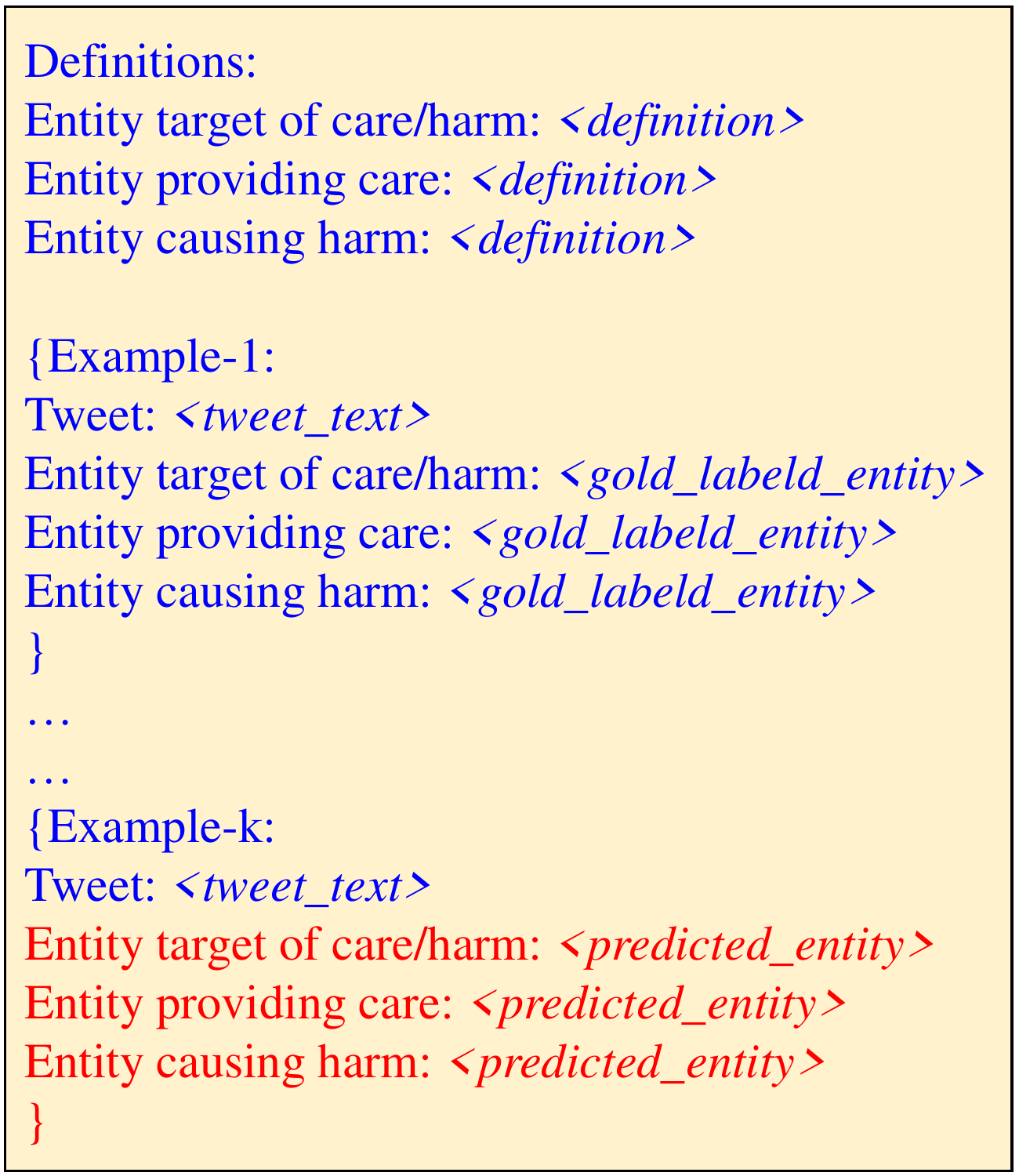}
  \caption{Prompt template for identification of entity and corresponding moral roles jointly in case of `Care/Harm'. The blue colored segment is input prompt and the red colored segment is the generated output by the LLMs. Corresponding prompt example can be seen in Appendix \ref{appx:prompt-examples}: Figure \ref{fig:mf-role-entity-direct-example}.}
  \label{fig:mf-role-entity-direct-template}
\end{figure}

In this approach, we propose a prompting method for the setting where the the entities are not pre-identified as described in Section \ref{sec:task-definition}. In this setting, the moral foundation is known for a tweet and the target entities in the tweets are not explicitly given. We create a prompt similar to a slot filling task where the LLMs have to fill the slots of moral roles with entities mentioned in the tweet. The prompt template is shown in Figure \ref{fig:mf-role-entity-direct-template}.

%% file: 5experiment.tex
\section{Experimental Evaluation}\label{sec:experimental-eval}
In this section first we discuss our experimental setting. Secondly, we discuss our proposed models' performance in morality frame identification.

\subsection{Experimental Settings}
\noindent\textbf{Large Language Model:} We use an open-source Large Language Model named GPT-J-6B \cite{gpt-j}. This is 6B parameters decoder only language model. We use top-k (k=5) sampling with temperature (=0.5) \cite{holtzman2019curious} as a decoding method for the language model. Note that, we do not update the parameters of the model in the in-context learning steps. For each of the test data point, we run the model with $5$ random seeds each generating $2$ outputs, hence, yielding 10 predictions for each data point. We take the majority voting among these predictions to get the predicted label.\\ 

\noindent\textbf{Ablation study:} We experiment with various numbers of training examples in the prompts. In this paper, we define number of shots or training examples $k$, as the number of examples used for training from each class related to a classification task. For moral foundation identification and moral roles identification of the pre-identified entities, we experiment with 0 to 5 shots. In the moral role identification method where entities are not pre-identified, we experiment with 0, 1, 3, 5, 7, 10 shots. Because of the limit in the number of tokens in the prompt we cannot experiment with more number of shots. In all of our prompting methods we provide the description of the expected labels as task instruction in the prompt. As a result, a zero-shot learning is feasible in our setting. We run all of the studies using the train and test set described in Section \ref{sec:dataset}.

\begin{table*}[ht!]
\centering
\resizebox{2\columnwidth}{!}{%
\begin{tabular}
{>{\arraybackslash}m{5cm}>{\centering\arraybackslash}m{1.5cm}>{\centering\arraybackslash}m{1.7cm}>{\centering\arraybackslash}m{1.7cm}>{\centering\arraybackslash}m{1.9cm}>{\centering\arraybackslash}m{1.7cm}>{\centering\arraybackslash}m{1.7cm}}

\toprule
    \multicolumn{1}{c}{} & \multicolumn{6}{c}{\textbf{Macro F1 score for various number of shots per class}}\\
    \cmidrule(lr){2-7}
    \textbf{Models} & \textbf{0-shot} & \textbf{1-shot} & \textbf{2-shots} & \textbf{3-shots} & \textbf{4-shots}  & \textbf{5-shots}\\
    \cmidrule(lr){1-1}\cmidrule(lr){2-7}
    One-Pass prompting for 5 classes & 6.24 & 24.19 & 29.80 & 30.63 & 39.49 & 43.56\\
    One-vs-all prompting           & 13.23 & 20.46 & 24.34 & 20.51 & 27.76 & 15.70\\
    \cmidrule(lr){1-7}
    RoBERTa (Parameters frozen)    & N/A & 7.61 (1.9) & 7.84 (2.3) & 8.1 (2.9) & 8.21 (3.1) & 8.0 (2.6)\\
    RoBERTa (Finetuned)             & N/A & 19.68 (7.3) & 33.22 (9.6) & 37.05 (5.8) & 38.78 (5.9) & 45.42 (6.6)\\
    \bottomrule
\end{tabular}}
\caption{Few-shot moral foundation identification results. Between the prompting-based methods, the one-pass prompting method is the best performing one. The one-pass prompting method outperforms parameters-frozen RoBERTa, but underperforms finetuned RoBERTa in few-shot training setup.}\label{tab:mf-identification}
\end{table*}

\begin{table}[ht!]
\begin{center}
\scalebox{0.7}{
\begin{tabular}{>{\arraybackslash}m{4cm}>{\centering\arraybackslash}m{1cm}>{\centering\arraybackslash}m{1cm}>{\centering\arraybackslash}m{1cm}>{\centering\arraybackslash}m{1.3cm}}
    \hline
    \textbf{Morals} & \textbf{Prec.} & \textbf{Rec.} & \textbf{F1} & \textbf{Support}\\
    \hline
    
    \textbf{Care/Harm}             &   31.82  &   70.00  &   43.75  &   20\\
    \textbf{Fairness/Cheating}     &   66.67  &   10.00  &   17.39  &   20\\
    \textbf{Loyalty/Betrayal}      &   31.43  &   55.00  &   40.00  &   20\\
    \textbf{Auth./Subversion}      &   87.50  &   35.00  &   50.00  &   20\\
    \textbf{Purity/Degradation}    &   100.0  &   50.00  &   66.67  &   20\\
    \hline
    \textbf{Accuracy}              &          &          &   44.00  &   100\\
    \textbf{Macro Average}         &   63.48  &   44.00  &   43.56  &   100\\
    \textbf{Weighted Average}      &   63.48  &   44.00  &   43.56  &   100\\
    \hline
\end{tabular}
}
\caption{Per class moral foundation classification results for one-pass prompting (using 5-shots per class).}
\label{tab:per-class-classification-results-moral-foundation}
\end{center}
\end{table}

\noindent\textbf{Baseline:} We compare our models' performance with a few-shot RoBERTa-based \cite{liu2019roberta} text classifier. For the identification of moral foundation in a tweet, we encode the tweet using RoBERTa where the embedding of the [CLS] token of the last layer is used as a representation of the text. This representation is used for moral foundation classification. For moral role identification of an entity in the tweet, we encode the tweet and the entity using two RoBERTa instances, and concatenate their representations to get a final representation. This concatenated representation is used for moral roles classification. Note that, the RoBERTa-based classifiers are trained with few-shot examples only as the prompting based methods. We run the RoBERTa-based classifiers 5 times using 5 random seeds and report the average result. 

\noindent\textbf{Implementation Infrastructure} We ran all of the experiments on a 4 core Intel(R) Core(TM) i5-7400 CPU @ 3.00GHz machine with 64GB RAM and two NVIDIA GeForce GTX 1080 Ti 11GB GDDR5X GPUs. GPT-J-6B was mounted using two GPUs. We used PyTorch library for all of the implementations.

\subsection{Results}
\noindent\textbf{Moral Foundation Identification:} In Table \ref{tab:mf-identification}, we show the results for moral foundation identification using our two proposed methods and few-shot RoBERTa. It can be seen that as the number of shots increases the performance improves in almost all of the cases. We also found that performance with RoBERTa is pretty bad with no gradient updates. But fine-tuning RoBERTa with few-shot examples provide reasonable performance. We found that the one-vs-all prompting technique underperforms the one-pass prompting technique, except in the zero-shot setting. Our intuition is that the language model is able to learn better when more contrastive examples are given which is the case in the one-pass method. Per class classification results for one-pass prompting using 5-shot examples per class are shown in Table \ref{tab:per-class-classification-results-moral-foundation}. However, the one-pass prompting technique outperforms the one-vs-all technique but underperforms few-shot RoBERTa with finetuning. It seems that without fine-tuning the subtle moral foundation identification is a difficult task for the LLMs.\\

\begin{table*}[ht!]
\centering

\resizebox{2.1\columnwidth}{!}{%
\begin{tabular}
{>{\arraybackslash}m{3.5cm}>{\arraybackslash}m{3.5cm}>{\centering\arraybackslash}m{2cm}>{\centering\arraybackslash}m{2cm}>{\centering\arraybackslash}m{2cm}>{\centering\arraybackslash}m{2cm}>{\centering\arraybackslash}m{2cm}}
\toprule
    \multicolumn{1}{c}{} & \multicolumn{1}{c}{} & \multicolumn{5}{c}{\textbf{Macro F1 score for various number of shots per class}} \\
    \cmidrule(lr){3-7}
    \textbf{Moral Foundations} & \textbf{Models} & \textbf{1-shot} & \textbf{2-shots} & \textbf{3-shots} & \textbf{4-shots}  & \textbf{5-shots}\\
    \cmidrule(lr){1-1}\cmidrule(lr){2-2}\cmidrule(lr){3-7}
    
    \multirow{3}{*}{\textbf{\makecell[l]{Care/Harm}}} & One-Pass Prompting    & 48.21 & 58.61 & 74.37 & 70.98 & 68.41\\
                                                      & 2-Steps Prompting   & 37.77 & 42.04 & 58.29 & 68.97 & 63.76\\
                                                      & RoBERTa (Finetuned) & 31.67 (13.4)  & 35.79 (13.2) & 35.35 (14.0) & 30.64 (14.0) & 43.83 (26.0)\\
    \cmidrule(lr){1-7}
    \multirow{3}{*}{\textbf{\makecell[l]{Fairness/Cheating}}}   & One-Pass Prompting    & 42.92 & 71.86 & 75.95 & 82.26 & 74.65\\
                                                                & 2-Steps Prompting   & 40.91 & 71.28 & 72.64 & 74.92 & 68.70\\
                                                                & RoBERTa (Finetuned) & 26.89 (11.9) & 46.16 (6.0) & 43.06 (3.6) & 35.61 (15.2) & 42.95 (12.9)\\
    \cmidrule(lr){1-7}                                                            
    \multirow{3}{*}{\textbf{\makecell[l]{Loyalty/Betrayal}}}    & One-Pass Prompting    & 35.56 & 36.40 & 35.24 & 45.10 & 41.27\\
                                                                & 2-Steps Prompting   & 30.39 & 38.69 & 32.32 & 38.82 & 25.83\\
                                                                & RoBERTa (Finetuned) & 21.29 (3.0) & 28.39 (7.1) & 24.14 (11.5) & 37.73 (1.7) & 36.57 (8.2)\\
    \cmidrule(lr){1-7}                                                            
    \multirow{3}{*}{\textbf{\makecell[l]{Authority/Subversion}}}    & One-Pass Prompting    & 19.17 & 31.69 & 29.35 & 34.76 & 36.12\\
                                                                    & 2-Steps Prompting   & 21.85 & 31.69 & 30.67 & 31.47 & 29.56\\
                                                                    & RoBERTa (Finetuned) & 11.77 (0) & 28.02 (11.6) & 23.31 (11.3) & 20.08 (10.5) & 24.64 (6.0)\\
    \cmidrule(lr){1-7}                                                                
    \multirow{3}{*}{\textbf{\makecell[l]{Purity/Degradation}}}  & One-Pass Prompting    & 41.28 & 46.91 & 66.67 & 69.04 & 61.84\\
                                                                & 2-Steps Prompting   & 40.51 & 41.66 & 43.08 & 47.65 & 45.89\\
                                                                & RoBERTa (Finetuned) & 31.59 (7.9) & 40.15 (5.7) & 30.80 (9.9) & 42.25 (10.8) & 56.57 (20.4)\\
    \bottomrule
\end{tabular}}
\caption{Few shot moral role identification performance comparison among models. The one-pass prompting method outperforms both 2-steps prompting method and finetuned RoBERTa in few-shot training setup.}\label{tab:moral-role}
\end{table*}

\noindent\textbf{Moral Role Identification for pre-identified entities:} In moral role identification, the assumption is that the moral foundation for each tweet is pre-identified. But the performance of all the models for the moral foundation identification task are not up to the mark as shown in Table \ref{tab:mf-identification}. So, in identification of moral roles we use the gold moral foundation labels instead of the predicted ones. 

In Table \ref{tab:moral-role}, we present the results for moral role identification using our proposed two methods along with the RoBERTa-based baseline. We omitted the results using zero-shot prompting as we found out that in moral role generation, zero-shot prompting of the LLM generates a lot of open-ended labels rather than the fixed moral role labels. It becomes difficult to parse these generations and map them to a moral role label using an automatic method. So we leave zero-shot prompting for moral role identification as a future work.

It can be seen in Table \ref{tab:moral-role} that both one-pass prompting and the two steps prompting methods outperform the RoBERTa baseline in moral role identification. It suggests that moral role identification is easier than moral foundation identification for LLMs. Note that, moral roles are micro structures of the morality frames and they are more focused towards entities and evident in text compared to subtle moral foundations. As a result it is easier for the LLMs to identify them. 

The two-steps prompting technique for moral roles identification underperforms the one-pass prompting approach although the task is broken down in two easier tasks. We found that in the first step of the task the model identifies polarity of sentiment towards entities with more than 70\% F1 score in the 4 shots and 5 shots settings. But it struggles in the second step where the model has to differentiate between two positive sentiments (e.g. `Entity target of care/harm' vs `Entity providing care') which is more difficult as the difference among positive sentiments is subtle. This finding is consistent with prior studies. For example, in previous work \citep{roy2021identifying} it was found that deep relational learning based model also struggles to differentiate among multiple positive sentiments. In the one-pass prompting technique, contrastive positive and negative examples are given in the prompt. As a result it might be easier for the LLMs to resonate. 

In moral role identification also the performance improves with the increase of number of shots for all of the models as shown in Table \ref{tab:moral-role}.\\

\noindent\textbf{Identification of entities and corresponding moral roles jointly:} In this setting, the model is expected to identify entities having the moral roles in a tweet. To evaluate the model's performance we measure in what percentage of time the predicted entity is matched with the actual entity\footnote{Entity matching procedure can be found in Appendix \ref{appx:entity-matching}} annotated by \citet{roy2021identifying} and in how many cases they are assigned to the correct entity role. We found out that the LLM hallucinates a lot when identifying entities and filling the entity role slots. Hallucination in LLMs is a common phenomena. When open ended text generation is expected but the language model generates some response that is not a part of the input text or not related to the input text, it is called hallucination \cite{ji2022survey}. Note that we don't encounter the problem of hallucination when generating labels for moral foundation and moral roles as the labels were well-defined in the prompt. But in entity identification task the model has to identify entities from a given text span which is open ended. Hence, it resulted in a higher rate of hallucination. 

However, The results for this task are shown in Table \ref{tab:entity-mf-joint-identification}. We can see in the table that as we increase the number of training examples (shots) the \% of correct entity and entity role identification improve although the performance is not up to the mark even with the highest number of shots (10). We also found out that \% of hallucination decreases as the number of shots increases. This findings imply that joint identification of entity and entity role is a much difficult task for the LLMs but as we increase the number of shots the LLMs are able to understand the task better.

\begin{table}[t!]
\centering
\resizebox{1\columnwidth}{!}{%
\begin{tabular}
{>{\centering\arraybackslash}m{1cm}>{\centering\arraybackslash}m{2.8cm}>{\centering\arraybackslash}m{2.3cm}>{\centering\arraybackslash}m{2.6cm}}

\toprule
    \textbf{No. of Shots} & \textbf{\% Correct Entity Identification} & \textbf{\% Hallucination} & \textbf{\% Correct Role Identification}\\
    \cmidrule(lr){1-1}\cmidrule(lr){2-4}
    1   & 43.80 & 21.69 & 33.97\\
    3   & 48.28 & 11.54 & 41.09\\
    5   & 48.68 & 9.58  & 43.71\\
    7   & 49.91 & 7.68  & 45.27\\
    10  & 51.39 & 5.95  & 46.88\\
    \bottomrule
\end{tabular}}
\caption{Correctness of joint identification of entity and corresponding moral roles using in-context learning. The LLM hallucinates from previous training examples in open-ended entity identification. The percentage of hallucination decreases and the percentage of correct entity and correct role identification increase with the increase of the number of shots in prompt.}\label{tab:entity-mf-joint-identification}
\end{table}

%% file: 6summary_future_works.tex
\section{Summary and Future Works}\label{sec:summary-future-works}
In this paper, we apply few-shot in-context learning for identification of one of the psycho-linguistic knowledge representation framework named Morality Frames. We proposed different prompting methods to perform the task. We found that in-context learning using a comparatively smaller language model (GPT-J-6B) does not perform well in identification of moral foundations that are very subtle. But it excels in moral roles identification of entities that are more evident in text. We believe there is a lot of scope for improvement, and this study will encourage the application of in-context learning in more Computational Social Science related tasks. Below we list a few future directions of this work.

\begin{itemize}
    \item \textbf{Prompt selection:} Appropriate prompt selection based on the test data point has been successfully applied in in-context learning in different NLP tasks \citep{han2022meet}. Implementation of a dynamic prompt selection technique in morality frame identification task may boost the performance.
    
    \item \textbf{Incorporation of context in prompt:} In complex concepts such as moral foundation \citep{haidt2004intuitive,haidt2007morality} and framing \citep{boydstun2014tracking}, to name a few, the social context and the speaker's demographics play an important role. Incorporating these information in prompts for LLMs can be an effective direction towards solving these problems.
    
    \item \textbf{Experiment with larger language models:} Larger language models such as GPT-3 \cite{brown2020language} use more parameters and are trained on diverse data. As a result, they could be more successful in capturing nuanced social concepts, and result in better performance.  
    
    \item \textbf{Experiment with long text:} Identification of complex concepts like framing and moral foundation have been studied in longer text (e.g. news articles) in previous works \citep{card.2015,fulgoni-etal-2016-empirical,field2018framing,royweakly}. How successful the pre-trained language models can be on these tasks in longer text such as, news articles, can be an interesting future work. 
    
\end{itemize}

%% file: limitations_ethical_statement.tex
\section*{Acknowledgements}
We are thankful to the anonymous reviewers for their insightful comments. This project was partially funded by NSF CAREER award IIS-2048001.

\section*{Limitations}
The limitations of this paper are as follows.

\begin{itemize}
    \item Previous study \cite{johnson2018classification} has shown that a single tweet may contain multiple moral foundations. Multiple labels were not considered in this work. It may be the case that language models are successful on identifying only one of the moral foundations in such multi-label data points.
    
    \item Usage of large language models are expensive as they are resource-heavy. Due to that we could not run the prompt-based methods multiple times to perform a statistical significance test on the results. This is a limitation of our work.
    
    \item Due to resource-constraint and no availability of an open-source version we could not run our proposed prompt-based models with state-of-the-art larger language models, such as GPT-3. The insights and results reported in this paper may have been different if a larger language model was used.
    
    \item LLMs are pretrained on a huge amount of human generated text. As a result, they may inherently contain many human biases \citep{brown2020language,blodgett2020language}. We did not consider any bias that can be incorporated by the LLMs in the morality frames identification task.
    
\end{itemize}

\section*{Ethics Statement}
In this paper, we do not propose any new dataset rather we only experiment with existing datasets which are, to the best of our knowledge, adequately cited. We provided all experimental details of our approaches and we believe the results reported in this paper are reproducible. Any result or tweet text presented in this paper are either results of a machine learning model or taken from an existing dataset. They don't represent the authors' or the funding agencies' views on this topic. 
As described in the limitations sections, inherent bias in the large language models are not taken into account in this paper while experimenting. So, we suggest not to deploy the proposed algorithms in a real life system without further investigation on bias and fairness.

%% file: 7appendix.tex
\section{Prompt Examples}\label{appx:prompt-examples}
\begin{figure*}[h]
  \centering
  \includegraphics[width=0.9\textwidth]{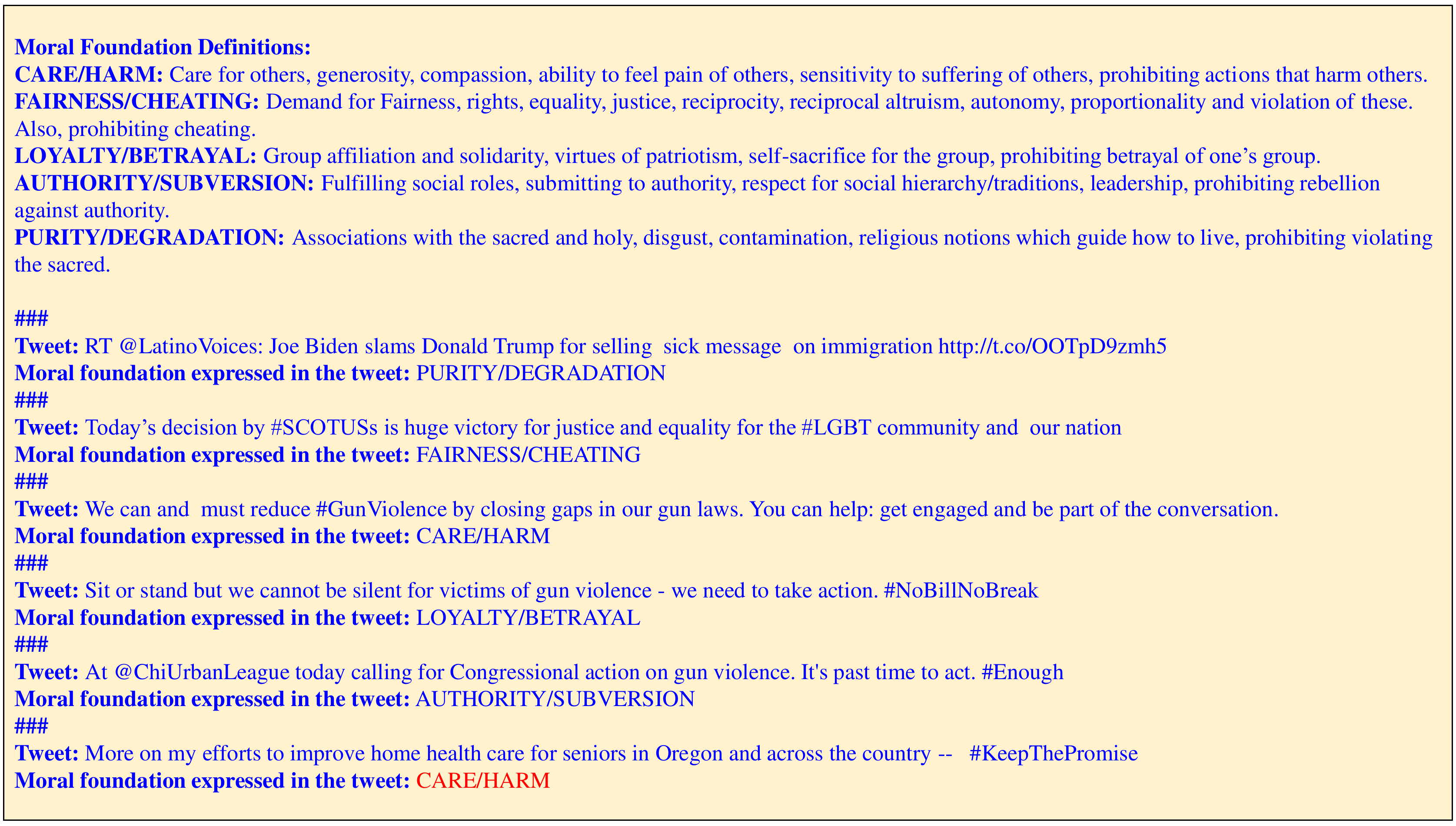}
  \caption{Prompt example for identification of moral foundation in one pass. The blue colored segment is input prompt and the red colored segment is the generated output by the LLMs.}
  \label{fig:mf-direct-example}
\end{figure*}

\begin{figure*}[h]
    \begin{center}
    \begin{subfigure}[t]{0.9\textwidth}
        \centering
        \includegraphics[width=1\textwidth]{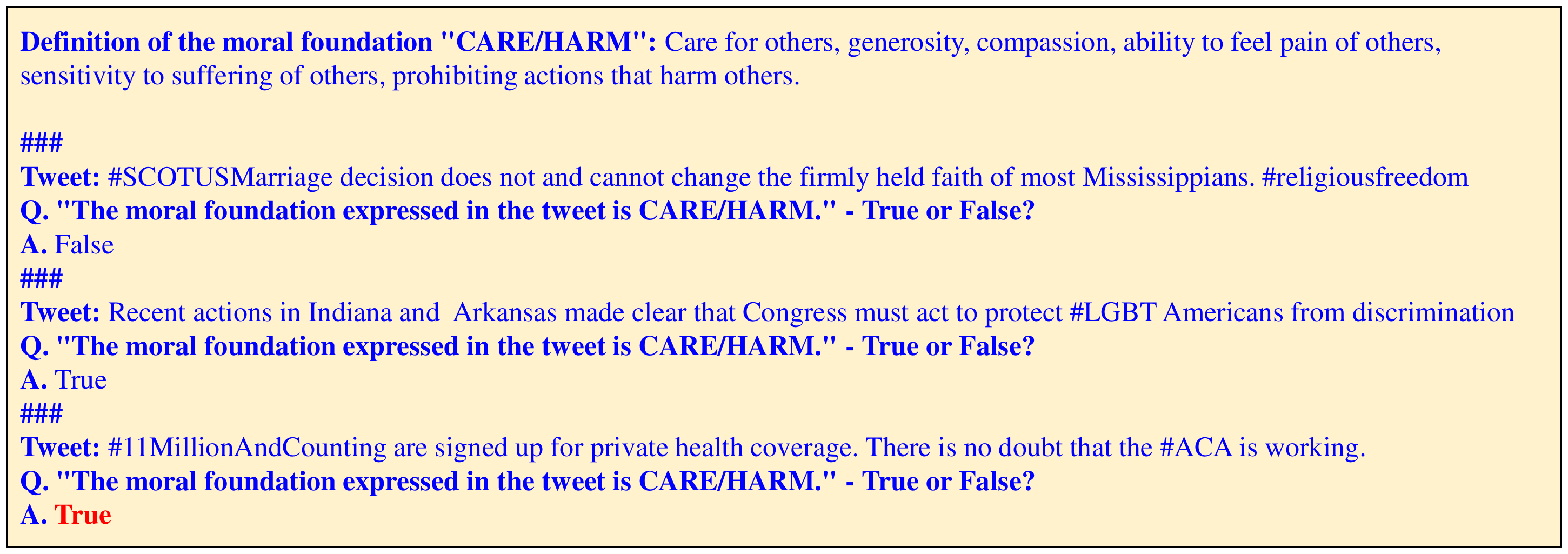}
        \caption{Prompt example for one-vs-all MF identification in case of `Care/Harm'.}
        \label{fig:step-1-example}
    \end{subfigure}
    \\
    \begin{subfigure}[t]{0.9\textwidth}
        \centering
        \includegraphics[width=1\textwidth]{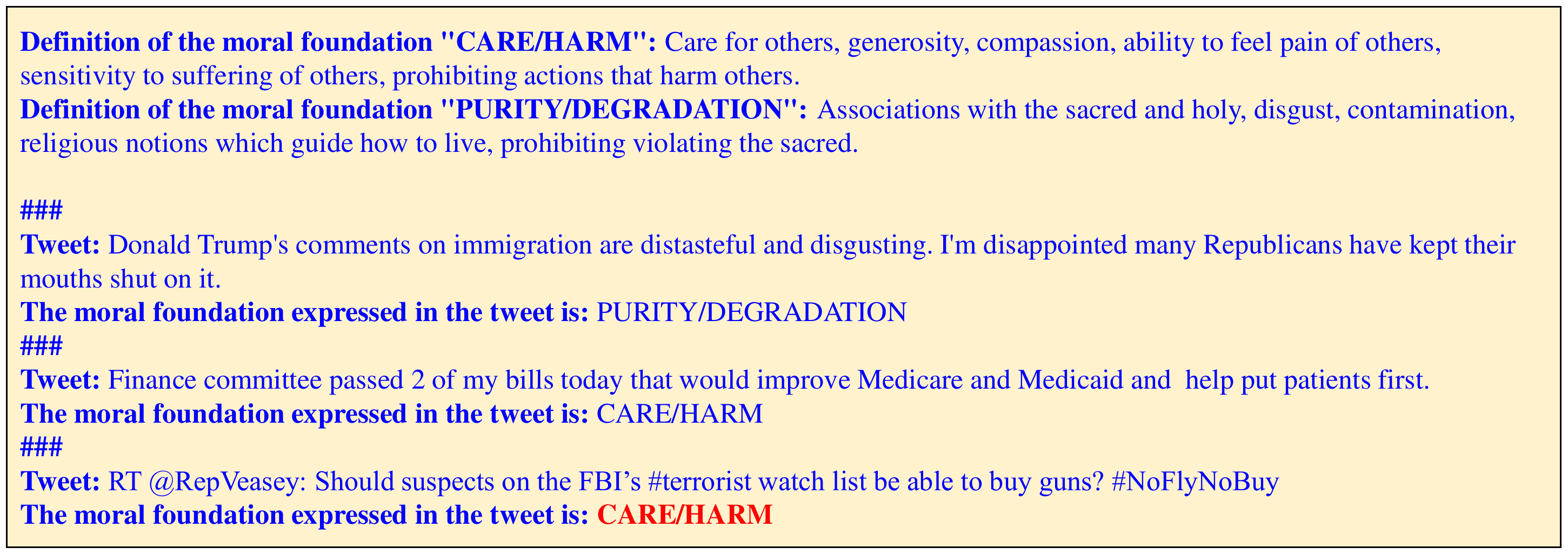}
        \caption{Prompt example for tie-breaking between two MFs. For example, between `Care/Harm' and `Purity/Degradation'.}
        \label{fig:step-2-example}
    \end{subfigure}
    \caption{Prompt examples for moral foundation identification technique in one-vs-all manner. The blue colored segments are input prompts and the red colored segments are the generated output by the LLMs.}
    \label{fig:mf-one-vs-all-example}
    \end{center}
\end{figure*}

\begin{figure*}[h]
  \centering
  \includegraphics[width=0.9\textwidth]{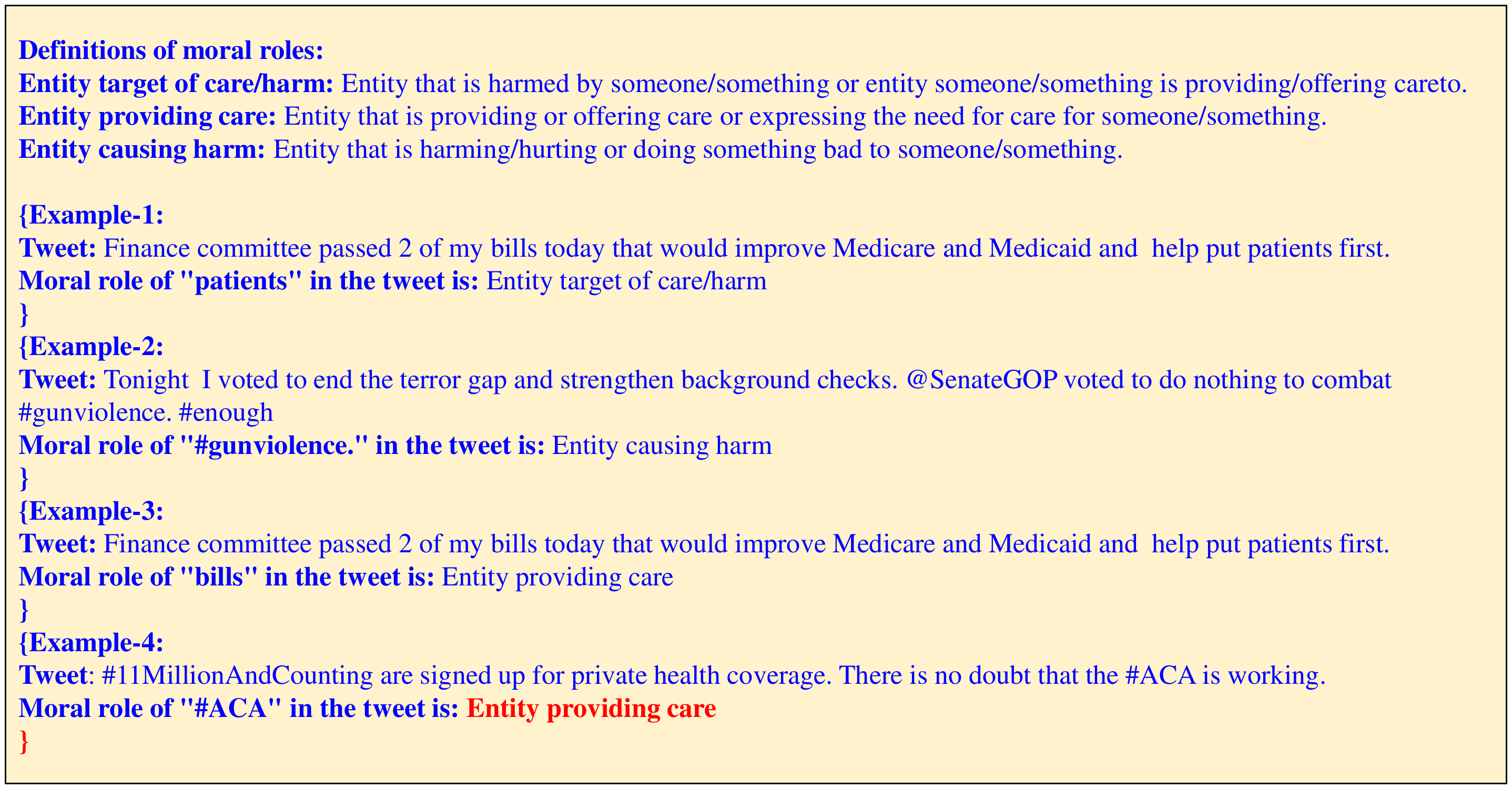}
  \caption{Prompt example for identification of moral role in one pass in case of `Care/Harm'. The blue colored segment is input prompt and the red colored segment is the generated output by the LLMs.}
  \label{fig:mf-role-direct-example}
\end{figure*}

\begin{figure*}[h]
    \begin{center}
    \begin{subfigure}[t]{0.9\textwidth}
        \centering
        \includegraphics[width=1\textwidth]{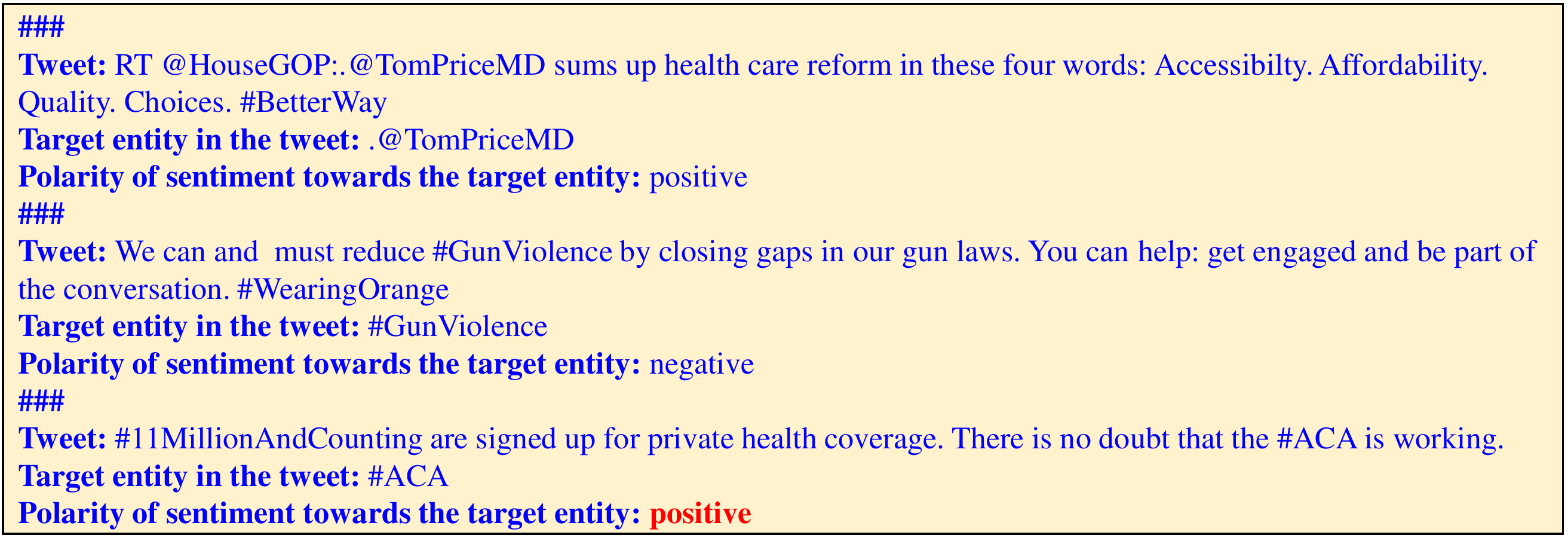}
        \caption{Step-1: Prompt example for identification of positive/negative sentiment towards entities.}
        \label{fig:mf-role-step-1-example}
    \end{subfigure}
    \\
    \begin{subfigure}[t]{0.9\textwidth}
        \centering
        \includegraphics[width=1\textwidth]{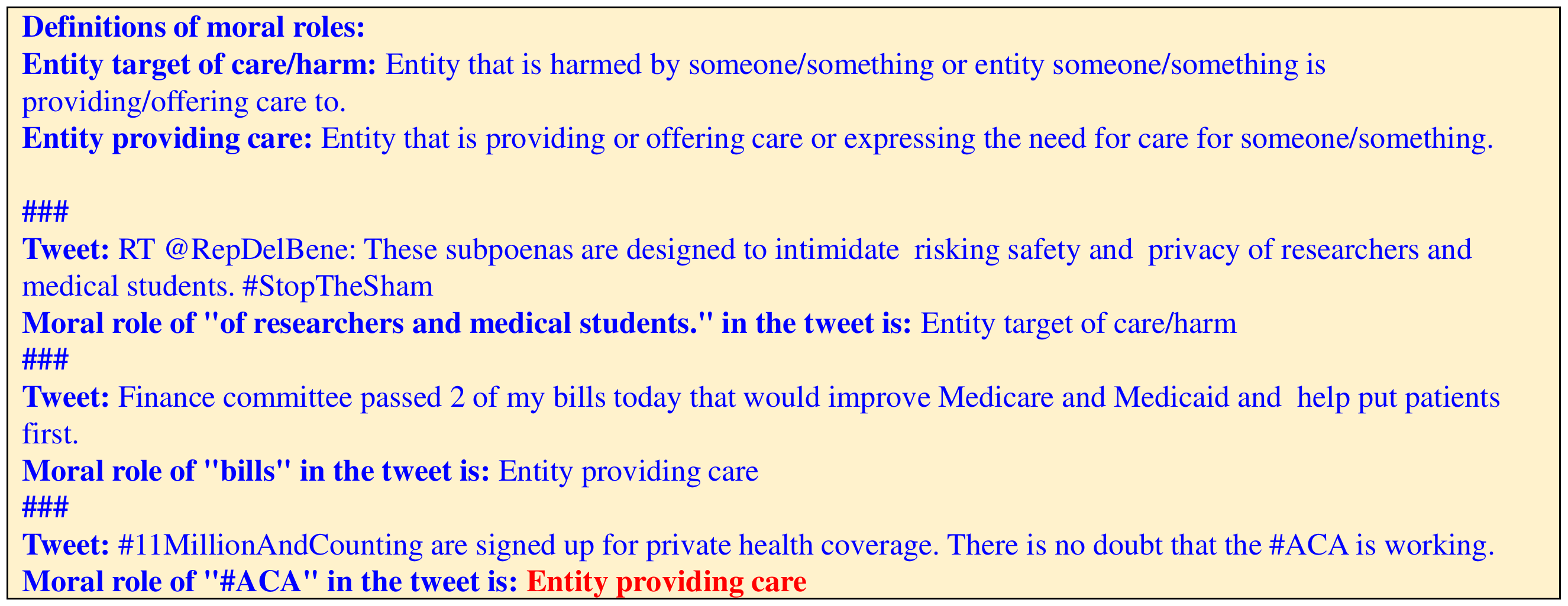}
        \caption{Step-2: Prompt example for differentiating among multiple positive moral roles in case of `Care/Harm'.}
        \label{fig:mf-role-step-2-example}
    \end{subfigure}
    \caption{Prompt examples for moral role identification by breaking it in two steps. The blue colored segments are input prompts and the red colored segments are the generated output by the LLMs.}
    \label{fig:mf-role-two-steps-example}
    \end{center}
\end{figure*}

\begin{figure*}[h]
  \centering
  \includegraphics[width=0.9\textwidth]{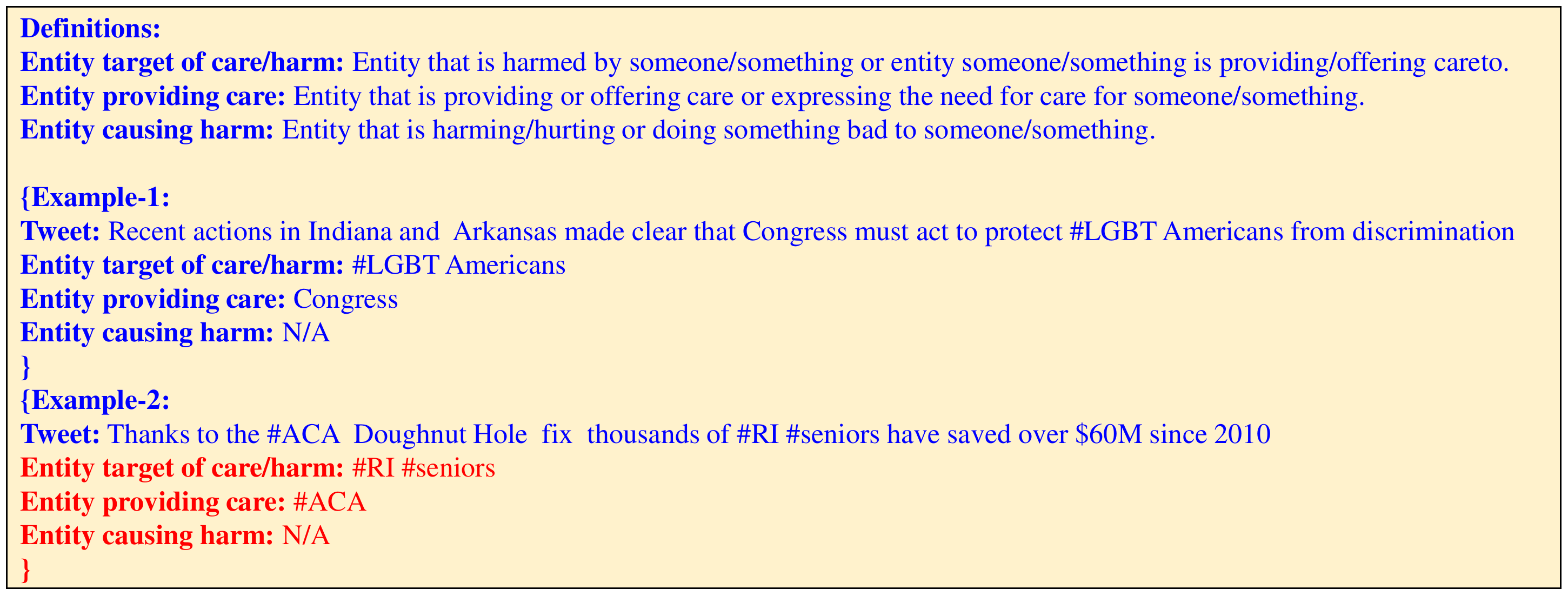}
  \caption{Prompt example for identification of entity and corresponding moral roles jointly in case of `Care/Harm'. The blue colored segment is input prompt and the red colored segment is the generated output by the LLMs.}
  \label{fig:mf-role-entity-direct-example}
\end{figure*}

The example of prompts for various in-context learning steps of our approach are shown in Figures \ref{fig:mf-direct-example}, \ref{fig:mf-one-vs-all-example}, \ref{fig:mf-role-direct-example}, \ref{fig:mf-role-two-steps-example}, \ref{fig:mf-role-entity-direct-example}.

\section{Entity Matching Procedure}\label{appx:entity-matching}
After obtaining the predicted entity labels from LLM, we first discard the entity labels that are not contained in the tweet text as these are irrelevant. Then, we check if any of the predicted entities are exactly matching the gold labels. In cases where it is not an exact match, we obtain a string-match score between the predicted entity and each of the gold label. If this score is beyond a certain threshold (set to 0.6) for a particular gold label, we map the predicted entity to that gold label. If the predicted entity is not exactly matching the gold label, and the score is lower than the threshold, then we assign 'N/A' label to that predicted entity.